\documentclass[runningheads]{llncs}
\usepackage[T1]{fontenc}
\usepackage{graphicx}
\usepackage{cite}
\usepackage{multirow}
\usepackage{hyperref}
\begin{document}
\title{Advanced Tumor Segmentation in PET/CT Imaging: A Training Strategy Study with nnU-Net for AutoPET III}
%
%
\author{Hussain Alasmawi}
%
%
\institute{Mohamed bin Zayed University of Artificial Intelligence \\
\email{hussain.alasmawi@mbzuai.ac.ae}}
\maketitle              
\begin{abstract}
Tumor segmentation in whole-body PET/CT imaging is crucial for precise disease evaluation and treatment planning. However, it remains challenging due to variability in lesion size, contrast, and anatomical distribution. Relying on manual segmentation makes the process time-consuming and prone to intra- and inter-observer variability. This work presents a whole-body tumor segmentation method developed for the AutoPET III challenge, where the goal is to build models that generalize across tracers and multi-center data. We employ the nnU-Net framework with a ResNet-based encoder as our baseline and systematically investigate the impact of training strategies, including intensity normalization, batch dice optimization, and data augmentation using CraveMix. Our experiments show that these strategies significantly influence model performance, particularly in reducing false positives and improving robustness to lesion variability. The best-performing configuration achieves a Dice score of up to 0.80 on the preliminary test phase, and our method ranked third in the AutoPET III challenge. The code is publicly available \href{https://github.com/HussainAlasmawi/AutoPet_Final}{here}.

\end{abstract}
\section{Introduction}

Cancer can arise in virtually any organ of the human body, leading to lesions distributed across diverse anatomical regions. Numerous approaches have been proposed to improve automatic tumor segmentation and enhance diagnostic accuracy. In this context, the AutoPET challenge \cite{tcia_fdg_pet_ct_lesions, gatidis2024autopet} has emerged as a prominent benchmark for whole-body tumor segmentation using PET/CT imaging. The third edition of the challenge focuses on developing models that generalize across multiple tracers and data from different medical centers. This manuscript presents our contribution to this task.

In summary, we employ the nnU-Net \cite{nnunet} framework with a ResNet-based encoder for tumor segmentation and systematically investigate the impact of training strategies, including intensity normalization, batch Dice optimization, and data augmentation using CraveMix \cite{cravemix}.

\section{Methods}
\subsection{Dataset}
The AutoPET-III dataset consists of whole-body PET/CT scans from patients diagnosed with various malignancies, including melanoma, lymphoma, prostate cancer, and lung cancer. The data were collected from two medical centers and comprise two primary cohorts: FDG and PSMA. The FDG cohort includes 501 patients with cancer and 513 negative controls. The PSMA cohort contains 537 cases with PSMA-avid tumor lesions and 60 cases without, including pre- and post-treatment scans of male patients.

\subsection{Network Architecture}
We employ the nnU-Net framework \cite{nnunet} as our baseline architecture for whole-body tumor segmentation. Specifically, we utilize a Residual Encoder UNet variant, which incorporates a ResNet-based encoder to improve feature representation and gradient flow during training. The nnU-Net framework automatically adapts key architectural and training configurations based on the dataset characteristics, making it a strong baseline for medical image segmentation tasks.

For our experiments, we adopt the Residual Encoder Large configuration and train the model using a patch size of 192×192×192 voxels. This configuration has been shown to provide improved performance in prior studies \cite{first,fabian2023}, particularly for large-scale 3D medical imaging tasks.

\subsection{Inference}
Given the large size of whole-body PET/CT volumes relative to the model input, we employ a sliding window inference strategy, as implemented in the nnU-Net framework, to generate full-resolution segmentation predictions. This approach enables efficient processing of high-resolution 3D images while preserving spatial consistency across patches. To improve prediction robustness, we leverage model ensembling by aggregating the outputs of models trained using 5-fold cross-validation, where final predictions are obtained by averaging the predicted probabilities across folds.

\section{Experiments \& Results}

We conduct a series of experiments using the nnU-Net framework, each evaluated with 5-fold cross-validation. All models are trained from scratch without the use of external datasets. The results are summarized in Table~\ref{tab:my-table}.

\begin{enumerate}
\item \textbf{Baseline}:  
We adopt the default nnU-Net configuration, with modifications limited to a patch size of $192 \times 192 \times 192$ voxels and intensity normalization using nnU-Net's standard CT normalization scheme for both CT and PET images.

\item \textbf{Z-score Normalization}: 
We replace the default PET normalization with Z-score normalization to evaluate its effect on segmentation performance.

\item \textbf{BraTS 2020 Strategy}: 
Inspired by the BraTS 2020 winning approach~\cite{brats2020}, we increase the batch size from 2 to 5 and employ batch Dice loss, which computes the Dice score across the entire batch rather than per individual sample. This strategy is designed to better handle variability in tumor size across patients.

\item \textbf{CraveMix Augmentation}: 
We incorporate CraveMix data augmentation~\cite{cravemix}, generating 350 synthetic samples per fold to improve model generalization.
\end{enumerate}

\begin{table}[ht]
\centering
\caption{Performance comparison of different training strategies using the nnU-Net framework. Results are reported as the average over 5-fold cross-validation and on the preliminary test phase, using Dice similarity coefficient (Dice), False Negatives (FN), and False Positives (FP).}
\label{tab:my-table}
\begin{tabular}{|l|ccc|ccc|}
\hline
\multirow{2}{*}{} & \multicolumn{3}{c|}{5-Fold Cross-Validation} & \multicolumn{3}{c|}{Preliminary Test Phase} \\ \cline{2-7}
 & Dice & FN & FP & Dice & FN & FP \\ \hline
Baseline   & 0.68 & 11.63 & 9.39 & 0.77 & 15.80 & 16.33 \\ \hline
BraTS2020  & 0.66 & 12.95 & 7.86 & 0.79 & 12.15 & 1.19  \\ \hline
CraveMix   & 0.67 & 11.26 & 11.06 & 0.78 & 11.12 & 16.16 \\ \hline
Z-score    & 0.68 & 13.19 & 8.90 & 0.80 & 18.74 & 4.09  \\ \hline
\end{tabular}
\end{table}

The results demonstrate that different training strategies lead to distinct performance trade-offs. In terms of cross-validation Dice score, both the baseline and Z-score normalization achieve the highest performance (0.68), indicating that the default nnU-Net normalization is already robust. However, Z-score normalization significantly reduces false positives in the test phase (FP = 4.09), albeit at the cost of an increased false negative rate (FN = 18.74).

The BraTS 2020 strategy yields a lower cross-validation Dice score (0.66) but improves generalization in the test phase, achieving a Dice score of 0.79. Notably, it substantially reduces false positives (FP = 1.19), suggesting improved robustness in handling variability in tumor size.

The CraveMix augmentation strategy provides moderate improvements in Dice score (0.67 in cross-validation and 0.78 in the test phase) but does not significantly reduce false positives or false negatives. This indicates that while augmentation increases data diversity, it does not necessarily translate to improved generalization in this setting.

Due to submission constraints, only two models were selected for the final submission. Based on preliminary test performance, the BraTS 2020 and Z-score normalization models were chosen.

\section{Conclusion}

In this work, we investigated the impact of training strategies on nnU-Net for whole-body tumor segmentation using PET/CT data from the AutoPET III dataset. Rather than modifying the model architecture, we focused on normalization schemes, batch-level optimization, and data augmentation. Our results demonstrate that these training choices play a critical role in model performance, leading to distinct trade-offs between Dice score, false positives, and false negatives. In particular, Z-score normalization significantly reduces false positives, while the BraTS 2020-inspired strategy improves generalization and robustness to lesion variability. These results emphasize the importance of training protocol design in nnU-Net-based segmentation systems and demonstrate that significant performance gains are possible without architectural changes. Our method achieved competitive performance with a third place in the AutoPET III challenge. =
%
%
%
%
\bibliographystyle{splncs04}
\bibliography{ref.bib}
\end{document}